%% file: main.tex
\documentclass{article}

\usepackage{arxiv}

\usepackage{natbib}
\usepackage[utf8]{inputenc}
\usepackage[T1]{fontenc}
\usepackage[utf8]{inputenc}
\usepackage{authblk}
\usepackage{hyperref}
\usepackage{url}
\usepackage{booktabs}
\usepackage{amsfonts}
\usepackage{nicefrac}
\usepackage{microtype}
\usepackage{xcolor}
\usepackage{enumitem}

\input{macros}

\title{\textbf{\sys: Scalable Model-Agnostic Data Valuation}}

\author[1]{Jiongli Zhu\textsuperscript{*}}
\author[1]{Parjanya Prajakta Prashant\textsuperscript{*}}
\author[1]{Alex Cloninger}
\author[1]{Babak Salimi}

\affil[1]{University of California San Diego}

\affil[ ]{\texttt{jiz143@ucsd.edu, pprashant@ucsd.edu, acloninger@ucsd.edu, bsalimi@ucsd.edu}}

\begin{document}
\maketitle

\begin{abstract}
Training data increasingly shapes not only model accuracy but also regulatory compliance and market valuation of AI assets. Yet existing valuation methods remain inadequate: \emph{model-based} techniques depend on a single fitted model and inherit its biases, while \emph{algorithm-based} approaches such as Data Shapley require costly retrainings at web scale. Recent Wasserstein-based model-agnostic methods rely on approximations that misrank examples relative to their true leave-one-out (LOO) utility. We introduce \sys, a scalable, model-agnostic valuation framework that assigns each example a \emph{distributional influence score}: its contribution to the Maximum Mean Discrepancy (MMD) between the empirical training distribution and a clean reference set. Unlike Wasserstein surrogates, our MMD-based influence admits a closed-form solution that faithfully approximates the exact LOO ranking within $O(1/N^2)$ error, requires no retraining, and naturally extends to conditional kernels for unified label- and feature-error detection. Moreover, \sys supports efficient online updates: when a new batch of size $\batchsize$ arrives, all scores can be updated in $O(\batchsize N)$ time, delivering up to $50\times$ speedup without compromising ranking quality. Empirical evaluations on noise, mislabeling, and poisoning benchmarks show that \sys consistently outperforms state-of-the-art model-, Shapley-, and Wasserstein-based baselines in both accuracy and runtime. We provide rigorous theoretical guarantees, including symmetry for reproducible rankings and density-separation for interpretable thresholds.
\end{abstract}

\renewcommand{\thefootnote}{}
\footnotetext{* These authors contributed equally to this work.}
\renewcommand{\thefootnote}{\arabic{footnote}}

\input{introduction}

\input{brief_literature_review}

\input{setup.tex}

\input{labels}

\input{experiments}

\input{conclusions}

\newpage

\bibliographystyle{plain}
\bibliography{main}

\input{appendix}

\end{document}

%% file: macros.tex
\usepackage[utf8]{inputenc}
\usepackage{amsmath,amssymb,amsthm}
\usepackage{algorithm}
\usepackage{algpseudocode}
\usepackage{algorithmicx}
\algrenewcommand\algorithmicrequire{\textbf{Input:}}
\algrenewcommand\algorithmicensure{\textbf{Output:}}
\usepackage{fullpage}
\usepackage{subcaption}
\usepackage{graphicx}
\usepackage{xspace}
\usepackage{bbm}
\usepackage{hyperref}
\usepackage{cleveref}
\usepackage{wrapfig}
\makeatletter
\newcommand*{\Relbarfill@}{\arrowfill@\Relbar\Relbar\Relbar}
\newcommand*{\xeq}[2][]{\ext@arrow 0055\Relbarfill@{#1}{#2}}
\makeatother
\usepackage{amsfonts}
\usepackage{mathtools}
\newtheorem{definition}{Definition}
\newtheorem{proposition}{Proposition}
\newtheorem{theorem}{Theorem}
\newtheorem{example}{Example}
\usepackage{amsmath,amssymb}

\definecolor{white}{rgb}{1,1,1}
\definecolor{black}{rgb}{0,0,0}
\definecolor{grey}{rgb}{0.7,0.7,0.7}
\definecolor{dgrey}{rgb}{0.5,0.5,0.5}
\definecolor{lightgrey}{rgb}{0.88,0.88,0.88}
\definecolor{lgrey}{rgb}{0.9,0.9,0.9}
\definecolor{llgrey}{rgb}{0.93,0.93,0.93}
\definecolor{lllgrey}{rgb}{0.96,0.96,0.96}
\definecolor{tableHeadGray}{rgb}{0.85,0.85,0.85}
\definecolor{oddRowGrey}{rgb}{0.95,0.95,0.95}
\definecolor{evenRowGrey}{rgb}{0.85,0.85,0.85}
\definecolor{yellow}{rgb}{1.0, 1.0, 0.0}
\definecolor{lightyellow}{rgb}{1.0, 1.0, 0.88}
\definecolor{selectiveyellow}{rgb}{1.0, 0.73, 0.0}
\definecolor{shadered}{rgb}{1,0.85,0.85}
\definecolor{red}{rgb}{1,0,0}
\definecolor{shadegreen}{rgb}{0.95,1,0.95}
\definecolor{green}{rgb}{0,1,0}
\definecolor{darkgreen}{rgb}{0,0.3,0}
\definecolor{shadeblue}{rgb}{0.95,0.95,1}
\definecolor{blue}{rgb}{0,0,1}
\definecolor{darkblue}{rgb}{0,0,0.5}
\definecolor{darkpurple}{rgb}{0.4,0,0.4}
\definecolor{darkdarkpurple}{rgb}{0.3,0,0.3}
\definecolor{eqhighlightcolor}{rgb}{0.7,0,0.0}

\newcommand{\dscifar}{CIFAR\mbox{-}10\xspace}
\newcommand{\dsstl}{STL\mbox{-}10\xspace}
\newcommand{\dsimdb}{IMDB\xspace}
\newcommand{\dsagnews}{AG News\xspace}

\newcommand{\MMD}{\ensuremath{\mathrm{MMD}}\xspace}
\newcommand{\MMDsquared}{\ensuremath{\mathrm{MMD}^2}\xspace}
\newcommand{\MCMD}{\ensuremath{\mathrm{MCMD}}\xspace}
\newcommand{\EMCMD}{\ensuremath{\mathrm{E\mbox{-}MCMD}}\xspace}

\newcommand{\E}{\ensuremath{\mathbb{E}}\xspace}
\newcommand{\IF}{\ensuremath{\mathrm{IF}}\xspace}
\newcommand{\sys}{\textsc{Kairos}\xspace}
\newcommand{\lava}{\textsc{Lava}\xspace}
\newcommand{\dataoob}{\textsc{DataOOB}\xspace}
\newcommand{\knnshapley}{\textsc{KNNShapley}\xspace}
\newcommand{\dvrl}{\textsc{DVRL}\xspace}

\newcommand{\batchsize}{m\xspace}
\newcommand{\AvgTrainK}{\mathcal{A}\xspace}
\newcommand{\AvgValK}{\mathcal{B}\xspace}

\newcommand{\FinalScore}{\mathcal{V}\xspace}
\newcommand{\Residual}{\mathcal{R}\xspace}

\newcommand{\fstar}{f^\star}
\newcommand{\eqsize}{\small}

%% file: introduction.tex
\vspace{-3.5mm}\section{Introduction}\label{sec:intro}
\vspace{-1.5mm}

Training data now dictates not only model accuracy but also regulatory risk and the market valuation of AI assets. Yet most valuation methods split into just two camps. \emph{Model-based} techniques, exemplified by influence functions~\citep{koh2017understanding} and TracIn~\citep{pruthi2020tracin}, probe a single trained model, so their scores inherit that model’s idiosyncrasies and can flip whenever the model or its hyper-parameters change~\citep{Karthikeyan2022InfluentialExamples}. At the other extreme, \emph{algorithm-based} methods such as Data Shapley~\citep{ghorbani2019data} average a point’s marginal contribution over many retrains; the combinatorial explosion makes them infeasible at web scale~\citep{ghorbani2019data,jia2019efficient}. Both strands therefore fall short of delivering valuations that are simultaneously stable and tractable for modern, billion-example datasets.

These methodological shortcomings collide with market trends and tightening regulatory scrutiny, driving demand for valuation techniques that remain reliable and computationally practical at scale. Curated text and image datasets now command eight-figure licensing fees—as exemplified by recent agreements between OpenAI and entities such as \emph{The Times} and Shutterstock~\citep{Reuters_Time_OpenAI_2024,Sherwood_Shutterstock_2024}.

Nevertheless, such curated datasets are expensive and limited in scope, forcing practitioners to rely on vast, noisy web crawls.
Concurrently, policymakers are increasingly mandating \emph{auditable, point-wise provenance}. Article 10 of the EU AI Act\footnote{Article 10 requires that high-risk systems use training data that are ``relevant, sufficiently representative, free of errors and complete,'' with substantial penalties for non-compliance; similar standards appear in the NIST AI RMF and FTC guidance~\citep{EU_AI_Act_2024,DLA_Piper_2025_EUAIAct}.} exemplifies these strict data-quality standards, which are echoed across multiple regulatory frameworks. Bridging the gap between commercial pressures and regulatory demands has spurred interest in new classes of \emph{model-agnostic} valuation techniques—including recent approaches like \lava~\citep{just2023lava,Kessler2025SAVA}—which aim to deliver scores that are simultaneously \emph{reliable}, \emph{scalable}, and \emph{robust}, without reliance on any specific trained model.

\begin{figure}[t]
  \centering
  \begin{subfigure}{0.24\textwidth}
    \centering
    \includegraphics[width=\linewidth]{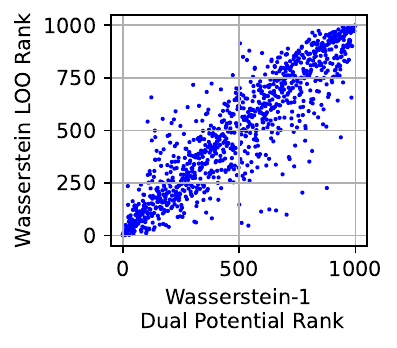}
    \vspace{-6mm}
    \caption{}
    \label{fig:intro:exact-dual}
  \end{subfigure}
  \hfill
  \begin{subfigure}{0.24\textwidth}
    \centering
    \includegraphics[width=\linewidth]{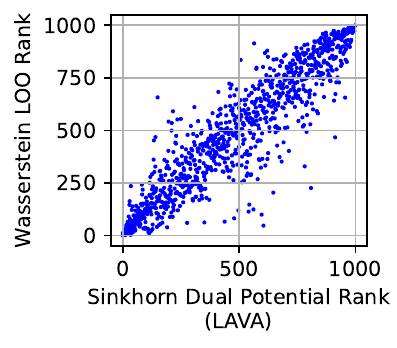}
    \vspace{-6mm}
    \caption{}
    \label{fig:intro:sinkhorn-dual}
  \end{subfigure}
  \hfill
  \begin{subfigure}{0.24\textwidth}
    \centering
    \includegraphics[width=\linewidth]{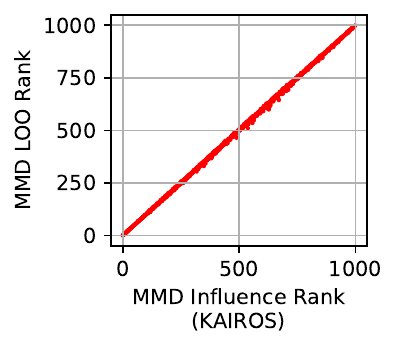}
    \vspace{-6mm}
    \caption{}
    \label{fig:intro:mmd}
  \end{subfigure}
  \hfill
  \begin{subfigure}{0.23\textwidth}
    \centering
    \includegraphics[width=\linewidth]{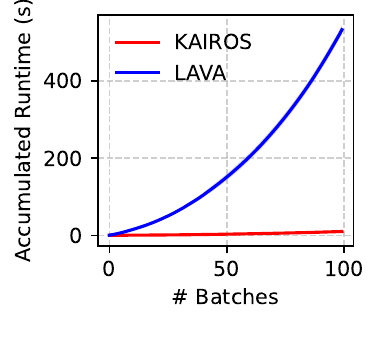}
    \vspace{-6mm}
    \caption{}
    \label{fig:intro:online-speed}
  \end{subfigure}
\vspace{-2mm}\caption{Comparison of Wasserstein- and MMD-based influence methods.
(\textbf{a}) Exact leave-one-out (LOO) ranks versus dual potentials from unregularised optimal transport.
(\textbf{b}) LOO ranks versus Sinkhorn dual potentials (\lava).
(\textbf{c}) LOO ranks versus MMD directional derivatives (\sys), which lie almost perfectly on the diagonal.
Ranking fidelity matters: among the top-100 points, \lava\ overlaps with the true LOO set only 60\%, while \sys\ achieves 99\% overlap—critical for downstream tasks such as data curation, marketplace pricing, and fairness auditing. (\textbf{d}) Online runtime: \sys\ scales linearly with the batch size and incurs minimal overhead, whereas \lava\ grows quadratically.}\vspace{-4mm}

  \label{fig:intro-example}
\end{figure}

In this work we introduce \sys, a scalable and reliable framework for \emph{model-agnostic} data valuation.
Each training example is assigned a \emph{distributional influence score}: the change it induces in a statistical divergence between the empirical distribution \(P\) and a trusted reference \(Q\).
To avoid the prohibitive cost of leave-one-out (LOO) retraining, we define this score as the \emph{directional derivative} of that divergence with respect to an infinitesimal up-weighting of the example.
For many popular divergences, including Wasserstein-1, computing this derivative exactly is infeasible.  Existing methods approximate it using Kantorovich dual potentials for the vanilla Wasserstein-1 distance or its faster, entropically-regularised \emph{Sinkhorn} variant, as in \lava~\citep{just2023lava}.  We show that, in general, these shortcuts produce rankings that drift from the true LOO ordering, over-valuing some points and under-valuing others (\Cref{fig:intro:exact-dual,fig:intro:sinkhorn-dual}).

\sys\ instead leverages the \emph{maximum mean discrepancy} (MMD), for which we show the directional derivative has a closed-form expression that
(i) exactly matches LOO rankings (\Cref{fig:intro:mmd});
(ii) can be computed in \(\mathcal{O}(\batchsize N)\) time for a mini-batch of size \(\batchsize\), enabling streaming-friendly performance that outpaces Wasserstein-based baselines (\Cref{fig:intro:online-speed}); and
(iii) naturally extends to task-specific similarity kernels, e.g.\ incorporating label information, while retaining these guarantees.  Our main contributions include:
\vspace{-2mm}
\begin{enumerate}[leftmargin=10pt]
  \item We derive a \emph{closed-form} influence function for MMD and rigorously prove that each \sys\ score exactly matches this influence, providing an accurate approximation of leave-one-out (LOO) valuations within \(\mathcal{O}(1/N^{2})\) error.

  \item We show that \sys\ requires no model training or iterative optimization. Computing influence scores for a mini-batch of size \(\batchsize\) takes only \(\mathcal{O}(\batchsize N)\) time and \(\mathcal{O}(N)\) memory, enabling efficient web-scale and streaming deployments.

  \item We establish that \sys\ satisfies the symmetry and density-separation axioms, ensuring fair rankings for points that contribute equally and clear separation of low- and high-quality data.

  \item The framework naturally extends to \emph{conditional} MMD kernels, allowing task-specific structure, such as label information, to be incorporated without sacrificing the above guarantees.

  \item Extensive experiments on label-noise, feature-noise, and back-door benchmarks demonstrate that \sys\ (i) flags corruptions earlier than model-, Shapley-, and Wasserstein-based baselines, (ii) preserves accuracy when pruning low-value points and sharply degrades it when removing high-value points, and (iii) runs up to \(50\times\) faster than the strongest prior method, \lava and \knnshapley.

  All code are available on \href{https://github.com/lodino/kairos}{\textcolor{blue}{GitHub}}.
\end{enumerate}

%% file: brief_literature_review.tex
\vspace{-2mm}
\section{Related Works}
\vspace{-3mm}
Data valuation methods broadly fall into three categories: model-based, algorithm-based, and model-agnostic approaches. Model-based methods include influence functions~\citep{koh2017understanding, grosse2023studying, choe2024your}, trajectory-based approaches~\citep{pruthi2020tracin, bae2024training, wang2024data}, and post-hoc kernel approximations like \textsc{Trak}~\citep{park2023trak} and \textsc{DaVinz}~\citep{wu2022davinz}. These methods produce model-specific valuations that may vary across different models trained on the same data. Algorithm-based methods, particularly Data Shapley~\citep{ghorbani2019data}, define value in terms of marginal contributions to specific learning algorithms, with various approximations developed to avoid the prohibitive cost of training $2^N$ models~\citep{jia2019towards, jia2019efficient, schoch2022cs, ghorbani2020distributional, wang2024data, kwon2021beta, wang2023data}. For bagging models, Data-OOB and 2D-OOB~\citep{kwon2023data, sun20242d} provide efficient alternatives using out-of-bag estimation. Algorithm-agnostic approaches like \textsc{Lava}~\citep{just2023lava} quantify contributions based on dataset distance but suffer from poor performance on label error detection~\citep{jiang2023opendataval}, computational complexity of $O(n^2)$, indeterministic approximations, and memory issues, which \textsc{Sava}~\citep{kessler2024sava} partially addresses through batching strategies for Wasserstein computation. A detailed literature review is provided in Appendix~\ref{section:related_work}.

%% file: setup.tex
\vspace{-3mm}
\section{Data Valuation via Distributional Influence}
\vspace{-4mm}

We cast data valuation in the framework of distributional sensitivity analysis. As in standard settings, we are given a training set \(D^{\text{train}} = \{(x_i, y_i)\}_{i=1}^{n_{\text{train}}}\) drawn i.i.d.\ from a noisy distribution \(Q\), and a validation set \(D^{\text{val}} = \{(x_i, y_i)\}_{i=1}^{n_{\text{val}}}\) from a clean target distribution \(P\). For a candidate point \((x, y)\), we ask: \textit{how much does it contribute to the distance between \(P\) and \(Q\)}? Concretely, we quantify the value as the change in distributional distance when the empirical training distribution is infinitesimally perturbed toward the Dirac mass at \((x, y)\). Since this influence depends only on \(P\), \(Q\), and a distributional distance \(d(P, Q)\), it defines an algorithm-agnostic valuation score.

The remainder of this section formalizes this idea (\S\ref{subsection:distributional_distances}), derives a closed-form score using \MMD (\S\ref{subsection:MMD}), extends it to labels using a conditional extension of \MMD (\S\ref{subsection:labels}), and presents an efficient algorithm for its computation (\S\ref{subsection:efficient}).

\vspace{-2mm}
\subsection{Distributional Distances}
\label{subsection:distributional_distances}
\vspace{-2mm}

In the absence of a learning algorithm, the distributional distance between \(P\) and \(Q\) serves as a surrogate for the train–validation risk gap. Several distance measures such as Total Variation, Wasserstein distance~\citep{Villani2003Topics, villani2008optimal}, and \MMD~\citep{gretton2012kernel} have been shown to upper bound the absolute difference between training and validation losses~\citep{ben2010theory, long2015learning, redko2017theoretical, courty2017joint}. We therefore value a training point by (the negative of) its contribution to the distance measure~\citep{just2023lava}. Specifically, we define this contribution by the influence function~\citep{law1986robust} of the distributional distance.

\begin{definition}[Distributional influence]
\label{defn:if}
Let \(d:\mathcal{M}\times\mathcal{M}\!\to\!\mathbb{R}\) be Gateaux-differentiable on the space \(\mathcal{M}\) of probability measures.  The influence of a point \(x\) is
\vspace{-1mm}{\eqsize\begin{equation}\label{eq:if}
\mathrm{IF}_d(x;P,Q)
= - \Bigl.\frac{d\bigl(P,(1-\varepsilon)Q+\varepsilon\delta_x\bigr)-d(P,Q)}{\varepsilon}\Bigr|_{\varepsilon\to0^+},
\end{equation}}\vspace{-1mm}
where \(\delta_x\) denotes the Dirac measure at \(x\).
\end{definition}

For finite samples, the influence function approximates the leave-one-out valuation with an error of \(O(1/n_{\text{train}}^2)\)~\citep{van2000asymptotic}.

\vspace{-2mm}
\paragraph{Choosing a distance.}
A distance \(d\) suitable for data valuation must (i) admit a tractable influence formula and (ii) upper bound the train–validation loss gap. \(f\)-divergences form an important family of distance measures, which include popular metrics such as Kullback--Leibler (KL) divergence~\citep{csiszar1967information, ali1966general}. However, they require density-ratio estimation (unstable in high dimensions~\citep{sugiyama2010density}) and are not well-defined whenever the support of \(P\) extends beyond that of \(Q\). Integral Probability Metrics (IPMs)~\citep{muller1997integral, Sriperumbudur2010Hilbert} avoid these pitfalls by not requiring density ratios in their definition or computation. Given two distributions \(P\) and \(Q\), the IPM is defined as \(d(P,Q)=\sup_{f\in\mathcal{F}}\bigl|\E_{P}[f]-\E_{Q}[f]\bigr|\), for a suitable function class \(\mathcal{F}\), where \(f\) is called the \emph{critic function} that aims to distinguish between $P$ and $Q$.
IPMs encompass various distance measures including Wasserstein-1~\citep{Villani2003Topics}, \MMD~\citep{gretton2012kernel}, and Total Variation Distance~\citep{muller1997integral}.

\vspace{-2mm}
\paragraph{IPM Influence decomposition.}
Substituting the IPM into the influence definition (Eq.~\eqref{eq:if}) gives
\begin{align}\eqsize
\label{eq:if-ipm-deriv}
\mathrm{IF}_{\text{IPM}}(x;P,Q)
&=-\Bigl.\frac{\sup_{f\in\mathcal{F}}\!\bigl[\E_{P}[f]-\E_{(1-\varepsilon)Q+\varepsilon\delta_x}[f]\bigr]-
           \sup_{f\in\mathcal{F}}\!\bigl[\E_{P}[f]-\E_{Q}[f]\bigr]}
          {\varepsilon}\Bigr|_{\varepsilon\to0^{+}} \\[2pt]
&=\underbrace{\Bigl.\frac{\E_{Q}[\fstar_{\varepsilon}-\fstar]-\E_{P}[\fstar_{\varepsilon}-\fstar]}{\varepsilon}\Bigr|_{\varepsilon\to0^{+}}}_{\text{(i) critic-shift term}}
+\underbrace{\bigl(\fstar(x)-\E_{Q}[\fstar]\bigr)}_{\text{(ii) point-gap term}}.
\label{eq:if-ipm}
\end{align}\vspace{-2mm}

Let $\fstar$ and $\fstar_{\varepsilon}$ denote the optimal critics before and after up-weighting a point $x$ by an infinitesimal mass~$\varepsilon$.
The \emph{critic-shift} term in the IPM influence function depends on $x$ and is generally intractable.  If the difference $(\fstar_{\varepsilon}-\fstar)$ decays as $O(\varepsilon^\alpha)$ for some $\alpha>1$, the critic-shift term vanishes and Equation~\eqref{eq:if-ipm} collapses to the simpler \emph{point-gap} term.
Crucially, this benign decay does \emph{not} hold for most IPMs—including Wasserstein-1 and total variation.  In the Wasserstein-1 case, $\fstar$ corresponds to a Kantorovich potential, which is \emph{non-unique}\citep{villani2008optimal}; discarding the critic-shift term therefore yields an influence that is non-deterministic and often unstable.

The state-of-the-art data-valuation method \lava~\citep{just2023lava} relies precisely on this simplification, retaining only the point-gap term for the Wasserstein-1 metric. This introduces two significant issues. (i) Since the dual critic is non-unique, the resulting influence values can vary arbitrarily, violating determinism. (ii) To enhance scalability, \lava\ replaces the exact Wasserstein-1 objective with its Sinkhorn-regularized counterpart; however, this introduces a bias of order \(O\bigl(d\,\nu\log(1/\nu)\bigr)\), where \(d\) is the data dimensionality and \(\nu\) is the regularization strength~\citep{genevay2019sample}. Correcting this bias is computationally expensive, and even as \(\nu \to 0\), the neglected critic-shift term remains unresolved, causing the resulting influence scores to deviate substantially from the true leave-one-out rankings (Figure~\ref{fig:intro:sinkhorn-dual}). Furthermore, the influence scores deviate from the true Sinkhorn leave-one-out rankings as well, demonstrating that the approximation error persists even with sinkhorn as the measure.

\subsection{Closed-form Influence via MMD}
\label{subsection:MMD}
We address these challenges by leveraging \MMD~\citep{gretton2012kernel}, an IPM that yields both closed-form distance calculations and elegant influence function derivations. \textit{Crucially, we derive a closed-form influence function that can be computed directly without requiring the expensive calculation of the distributional distance itself.} In this section, we focus on the marginal distribution of features, deferring the incorporation of label information to subsection~\ref{subsection:labels}.

Given two distributions $P$ and $Q$, the \MMD distance between them is defined as:
\begin{equation}\eqsize
\MMD(P, Q) = \sup_{\|f\|_{\mathcal{H}} \le 1} \bigl(\E_{x \sim P}[f(x)] - \E_{x \sim Q}[f(x)]\bigr)
= \|\mu_P - \mu_Q\|_{\mathcal{H}},
\end{equation}
where \(\mu_P = \E_{x\sim P}[\phi(x)]\) and \(\mu_Q = \E_{x\sim Q}[\phi(x)]\) are the kernel mean embeddings in the RKHS \(\mathcal{H}\), and the kernel \(k(x,x') = \langle\phi(x),\phi(x')\rangle_{\mathcal{H}}\)~\citep{gretton2012kernel}. While computing \(\|\mu_P - \mu_Q\|_{\mathcal{H}}\) directly can be challenging, the squared distance admits a closed-form expression via the kernel trick:
\begin{equation}\eqsize
\MMDsquared(P,Q) = \mathbb{E}_{x,x'\sim P}[k(x,x')] + \mathbb{E}_{x,x'\sim Q}[k(x,x')] - 2 \mathbb{E}_{x\sim P,\, x'\sim Q}[k(x,x')].
\end{equation}
This enables a closed-form computation of the influence with respect to \MMD via the chain rule.

\begin{proposition}
\label{proposition:mmd_influence}
The influence function for \MMD as the distance metric is, up to additive and positive multiplicative constants, given by
\begin{equation}\eqsize
\mathrm{IF}_{\mathrm{MMD}}(x; P, Q) = \mathbb{E}_{x' \sim P}[k(x', x)] - \mathbb{E}_{x' \sim Q}[k(x', x)].
\end{equation}
\end{proposition}

The full derivation is provided in Appendix~\ref{app:mmd_influence_proof}. Note that for downstream tasks such as feature error detection, backdoor attack identification, or ranking the most and least valuable points, only the relative rankings are important~\citep{kwon2021beta, wang2023data}; hence, the additive and multiplicative constants in the influence function do not affect the outcome. Henceforth, we use \(\IF(\cdot)\) to denote the rescaled version of \(\mathrm{IF}_{\mathrm{MMD}}(\cdot; P, Q)\), omitting \(P\) and \(Q\) for brevity. For a training point \( x_i \in D^{\text{train}} \), the unbiased finite-sample estimate of its \MMD-based influence is given by:
\begin{equation}\eqsize
\label{equation:empirical_if}
\widehat{\mathrm{IF}}(x_i) = \frac{1}{n_{\text{val}}} \sum_{j=1}^{n_{\text{val}}} k(x_j^{\text{val}}, x_i)
- \frac{1}{n_{\text{train}} - 1} \sum_{\substack{j=1, j \ne i}}^{n_{\text{train}}} k(x_j^{\text{train}}, x_i),
\end{equation}
where the first term approximates \(\mathbb{E}_{x' \sim P}[k(x', x_i)]\) using validation data and the second term approximates \(\mathbb{E}_{x' \sim Q}[k(x', x_i)]\) using the training data excluding \(x_i\).

\paragraph{Properties.}
\label{subsection:Properties}
Beyond closely approximating leave-one-out scores~\citep{van2000asymptotic}, our \MMD-based influence enjoys two structural guarantees that are critical in practice:
\emph{(i)~Symmetry} ensures that, for finite samples, points making the same marginal contribution to the \MMD receive identical influence, yielding fair rankings;
\emph{(ii)~Density-separation} guarantees the existence of a single, global threshold that cleanly partitions regions where the validation distribution dominates (\(P>Q\)) from those where the training distribution dominates (\(Q>P\)), enabling high-accuracy in detecting noisy samples and backdoor attacks.

\begin{proposition}[Symmetry]
\label{proposition:symmetry}
Let $D^{\text{train}}$ and $D^{\text{val}}$ be finite samples from distributions $Q$ and $P$, respectively.
If for all subsets $S \subseteq D^{\text{train}} \setminus \{x_i, x_j\}$,
{\eqsize\[\widehat{\MMD}(D^{\text{val}}, S \cup \{x_i\}) - \widehat{\MMD}(D^{\text{val}}, S) = \widehat{\MMD}(D^{\text{val}}, S \cup \{x_j\}) - \widehat{\MMD}(D^{\text{val}}, S)\]}
then $\widehat{\mathrm{IF}}(x_i) = \widehat{\mathrm{IF}}(x_j)$.
\end{proposition}

See Appendix~\ref{app:symmetry_proof} for proof.

\begin{proposition}[Density Separation]\label{prop:separation}
Let $P$ and $Q$ be two probability distributions on $\mathcal{X}\subseteq\mathbb{R}^n$. For any $\epsilon>0$ and $r>0$, there exists a Gaussian isotropic kernel $k$ such that for \( \mathrm{IF}(x) \;=\;\mathbb{E}_{x'\sim P}\bigl[k(x,x')\bigr]\;-\;\mathbb{E}_{x'\sim Q}\bigl[k(x,x')\bigr]\),
the following holds
{\eqsize\[
\begin{cases}
\mathrm{IF}(x)>0, & \text{if }P(x')-Q(x')\ge\epsilon\ \forall\,x'\in B(x,r),\\
\mathrm{IF}(x)<0, & \text{if }P(x')-Q(x')\le-\epsilon\ \forall\,x'\in B(x,r),
\end{cases}
\]}
where $B(x,r)=\{x':\|x'-x\|<r\}$.
\label{proposition:threshold_separation}
\end{proposition}
In Appendix~\ref{app:threshold_separation_proof} we give a full proof of Proposition~\ref{prop:separation}, and we also present a real-world example illustrating its practical effect, where the \MMD‐based influence scores for clean versus corrupted points are perfectly split by a near-zero threshold, whereas the Wasserstein‐based scores exhibit substantial overlap and admit no such clean cutoff.

%% file: labels.tex
\subsection{Capturing Feature–Label Correlations with MCMD}
\label{subsection:labels}

Marginal \MMD\ over \(P(X)\) and \(Q(X)\) effectively detects covariate shift but overlooks label-specific corruptions—such as flipped labels, back-door triggers, or concept drift—that alter \(P(Y\!\mid\!X)\) without changing \(P(X)\).
Directly operating on the joint distribution \((X,Y)\) using a product kernel often \emph{reduces} sensitivity to label noise: in high-dimensional feature spaces, the geometry of \(X\) dominates the kernel distances, causing small perturbations in labels \(Y\) to barely shift the joint embedding, and diminishing the test's power as dimensionality grows~\citep{ramdas2015adaptivity,reddi2015linear}.
We therefore retain marginal \MMD\ for detecting feature-level anomalies, and augment it with the expected value of Maximum Conditional Mean Discrepancy (\MCMD)~\citep{ren2016conditional,park2020measure}, a conditional extension of \MMD\ denoted as \EMCMD.
This hybrid criterion enables \sys\ to identify both covariate and label anomalies within a unified, kernel-based theoretical framework.

\begin{definition}[Maximum Conditional Mean Discrepancy (\MCMD)~\citep{ren2016conditional, park2020measure}]
The \MCMD between conditional distributions $P(Y|X)$ and $Q(Y|X)$ at point $x$ is:
{\eqsize\[
\MCMD_{P,Q}(x) := \|\mu_{Y|X}^P(x) - \mu_{Y|X}^Q(x)\|_{\mathcal{H}_\mathcal{Y}},
\]}
where $\mu_{Y|X}^P(x) = \int \phi(y)\, dP(y|x)$ and $\mu_{Y|X}^Q(x) = \int \phi(y)\, dQ(y|x)$ are known as the conditional kernel mean embeddings.
\end{definition}

To aggregate this measure across covariate space, we take the expectation of \(\MCMD(\cdot)\) over the training distribution i.e. $\EMCMD(P,Q) = \mathbb{E}_{x \sim Q}[\mathrm{MCMD}_{P,Q}(x)]$. We derive the influence for \EMCMD via Definition~\ref{defn:if}.

\begin{proposition}
\label{proposition:emcmd_influence}
The influence function for \EMCMD as the distance metric is, up to additive and positive multiplicative constants, given by
\begin{equation}\eqsize
\IF_{\EMCMD}(x,y; P, Q) = -\|\mu_{Y|X}^P(x) - \phi(y)\|_{\mathcal{H}_\mathcal{Y}}
\end{equation}
\end{proposition}

The full derivation is provided in Appendix~\ref{app:emcmd-influence-proof}. Henceforth, we use \(\IF_{cond}(x,y)\) to denote the rescaled version of \(\mathrm{IF}_{\EMCMD}((x,y); P, Q)\), omitting \(P\) and \(Q\) for brevity. We can simplify this expression when $Y$ is categorical ($Y \in \{0,1,\ldots,C-1\}$). For categorical labels, consider the feature map \(\phi(y) = e_y \in \mathbb{R}^C\) and the kernel \(k(y,y') = \mathbbm{1}\{y=y'\}\) where $e_y$ is the $y$-th standard basis vector. For this kernel and mapping function, we have \(\mu_{Y|X}^P(x) = [P(0|x), \ldots, P(C-1|x)]^T\) which is simply the probability vector for each class conditioned on $x$. Therefore:
\begin{equation}\eqsize
\IF_{\text{cond}}(x,y) = -\|\mu_{Y|X}^P(x) - e_y\|_2 = -\sqrt{\sum_{y' \neq y} P(y'|x)^2 + (P(y|x) - 1)^2}.
\end{equation}
For the finite sample case, $P(Y|X)$ can be estimated by a classifier trained on $D^{\text{val}}$ that returns $\hat{P}(y|x)$. The finite sample estimator for the conditional influence is:
\begin{equation}\eqsize
\label{equation:empirical_if_cond}
\widehat{\IF}_{\text{cond}}(x,y) = -\sqrt{\sum_{y' \neq y} \hat{P}(y'|x)^2 + (\hat{P}(y|x) - 1)^2}.
\end{equation}
\paragraph{Combined Net Distance and Influence.}
Finally, we integrate both marginal and conditional discrepancies into a single ``net’’ distance:
{\eqsize\begin{equation}\label{eq:net-dist}
  d_{\text{net}}(P,Q)
  \;=\;
  (1-\lambda)\;\MMD(P_X,Q_X)
  \;+\;\lambda\;\EMCMD(P,Q),
\end{equation}}
where $\lambda>0$ balances the two terms.  The overall influence of $(x,y)$ on $d_{\text{net}}$ is simply
{\eqsize\begin{equation}\label{eq:net-if}
  \IF_{\text{net}}(x,y)
  \;=\;
  (1-\lambda)\;\IF\bigl(x\bigr)
  \;+\;
  \lambda\;\IF_{\text{cond}}(x,y).
\end{equation}}

\input{theory}

\subsection{Batch Computation and Streaming Updates}
\label{subsection:efficient}

In this section we describe how \sys\ keeps influence estimates up to date in continuously evolving datasets. Modern ML pipelines, such as language models trained on fresh web crawls or recommender systems processing new user interactions, require efficient incremental data valuation rather than expensive recomputation from scratch. \sys\ tackles this by first running an \emph{offline initializer} that scans a static training–validation split once, caching the per-point kernel means and residual needed for its closed-form score; after this one-time pass, an \emph{incremental updater} processes each incoming mini-batch, assigns influence scores to the new points, and adjusts the cached statistics of existing points using only the kernel interactions introduced by the newcomers. This design maintains accurate valuations in a streaming setting while avoiding quadratic recomputation.

\paragraph{Offline Algorithm.}
In the standard offline setting, we assume access to the entire dataset and a classifier trained on $D^{\text{val}}$ that provides predicted probability vectors $\hat{y}_i^{\text{train}} = \hat{P}(y|x_i)$ for training points. For each training point $(x_i, y_i) \in D^{\text{train}}$, we precompute three key quantities:
\begin{equation}\eqsize
\label{equation:ai_bi_ri}
\AvgTrainK_i = \frac{1}{n_{\text{train}} - 1} \sum_{\substack{j=1 \\ j \ne i}}^{n_{\text{train}}} k(x_j^{\text{train}}, x_i), \quad
\AvgValK_i = \frac{1}{n_{\text{val}}} \sum_{j=1}^{n_{\text{val}}} k(x_j^{\text{val}}, x_i), \quad
\Residual_i = \|y_i^{\text{train}} - \hat{y}_i^{\text{train}}\|_2
\end{equation}
Using these precomputed values, the feature influence is $\widehat{\IF}(x_i) = \AvgValK_i - \AvgTrainK_i$ (from Equation~\eqref{equation:empirical_if}) and the conditional influence is $\widehat{\IF}_{\text{cond}}(x_i, y_i) = R_i$ (from Equation~\eqref{equation:empirical_if_cond}). Therefore, the final net influence score is:
\begin{equation}\eqsize
\widehat{\IF}_{\text{net}}(x_i, y_i) = (1-\lambda)(\AvgValK_i - \AvgTrainK_i) + \lambda \Residual_i
\end{equation}
The time complexity for this algorithm is $O(n_{\text{train}}^2 + n_{\text{train}} \cdot n_{\text{val}})$, which simplifies to $O(n_{\text{train}}^2)$ since typically $n_{\text{train}} > n_{\text{val}}$.

\paragraph{Online Algorithm.}
In the online setting, we process data in batches. At time $t$, we receive a new batch of size $\batchsize$, bringing the total dataset size to $n_{t+1} = n_t + \batchsize$. We update the influence scores for both existing points and the new batch. While a naive approach of recomputing everything from scratch would require $O(n_{t+1}^2)$ time, our elegant influence expression allows computation in $O(n_t \cdot \batchsize + \batchsize^2)$.

For existing points $(x_i, y_i)$ where $i \leq n_t$, the terms $\AvgValK_i$ and $\Residual_i$ remain unchanged. The training kernel average $\AvgTrainK_i$ can be efficiently updated as:
{\eqsize\[
\AvgTrainK_i^{(t+1)} = \frac{1}{n_{t+1}-1} \left((n_t - 1) \cdot \AvgTrainK_i^{(t)} + \sum_{j=1}^{\batchsize} k(x_{n_t+j}^{\text{train}}, x_i)\right)
\]}
Note that this update requires computing only $\batchsize$ new kernel evaluations per existing point, not $n_{t+1}$. For new points $(x_{n_t+i}, y_{n_t+i})$ where $i = 1, \ldots, \batchsize$, we compute all three quantities from scratch using Equation~\eqref{equation:ai_bi_ri}. The final influence scores are then computed as $\widehat{\IF}_{\text{net}}(x_i, y_i) = \lambda(\AvgValK_i - \AvgTrainK_i) + (1-\lambda)\Residual_i$ for all points. This update procedure runs in $O(n_t \cdot \batchsize)$ for the old data and $O( \batchsize^2)$ for the new batch resulting in total complexity of $O(n_t \cdot \batchsize + \batchsize^2)$. Detailed algorithm provided in Appendix~\ref{appendix:algorithm-online}.

%% file: theory.tex
\paragraph{Generalization Error Bound.}

We establish a theoretical link between the net discrepancy \(d_{\text{net}}\) and downstream model performance. Under mild regularity assumptions, the expected train–validation loss gap is bounded above by the sum of the marginal \MMD\ and conditional \EMCMD. Consequently, pruning or down-weighting points with large influence scores provably tightens an out-of-distribution error bound for \emph{any} learning algorithm. A concise statement follows, with the complete theorem and proof in Appendix~\ref{app:generalization-bound-proof}.

\begin{theorem}[Bounding transfer loss (simplified)]\label{theorem:generalization-bound}
Let $(\mathcal{X},d_{\mathcal{X}})$ and $(\mathcal{Y},d_{\mathcal{Y}})$ be compact metric spaces and $\mathcal{Z}=\mathcal{X}\times\mathcal{Y}$. Let $L:\mathcal{Z}\to\mathbb{R}$ be a continuous loss function.
Then, for some constant c,

{\eqsize\[
\mathbb{E}_{(x,y)\sim Q}[L(x,y)
] \leq
\mathbb{E}_{(x,y)\sim P}[L(x,y)] +
c
\Bigl(\,
\MMD(P_{X},Q_{X})
+
\mathbb{E}_{x\sim Q_{X}}\bigl[\mathrm{MCMD}_{P,Q}(x)\bigr]
\Bigr).
\]}

\end{theorem}

%% file: experiments.tex
\section{Experiments}\label{sec:exp}
\vspace{-3mm}
We implement \sys as a custom data valuation method in the OpenDataVal benchmark~\cite{jiang2023opendataval}, a benchmark for evaluating data valuation methods.
We evaluate all methods on three key applications: (1) detecting noisy data, mislabels, and mixture of them, (2) detecting malicious data injected by data poisoning attacks, and (3) data pruning by removing data with the lowest values. We also test removing data with the highest values to thoroughly examine the effectiveness of all data values, including low and high.
In addition, we test the runtime of methods with varying data sizes, in offline and online settings.
All our experiments are conducted on a single machine equipped with an Apple M1 chip, 8 cores, and 16 GB of RAM.
Experiments are repeated five times with different random seeds, and we report the mean (colored regions denote the standard deviations).
The code is shared on \href{https://github.com/lodino/kairos}{\textcolor{blue}{GitHub}}.

\vspace{-3mm}
\paragraph{Datasets.}

We evaluate on four widely used datasets, including \dscifar~\cite{krizhevsky2009learning}, \dsstl~\cite{coates2011analysis}, \dsimdb~\cite{maas2011learning}, and \dsagnews~\cite{zhang2015character}, to cover both image and text modalities.
In most experiments, we simulate limited clean-data availability by using 10000 noisy training examples and 300 clean validation examples, with a held-out test set of 10000 clean samples.
For the smaller \dsstl dataset, we scale down to 3700 training, 300 validation, and 1000 test examples.

\vspace{-3mm}
\paragraph{Baselines and Hyper-parameters.}
We compare \sys with four state-of-the-art data valuation methods with different mechanisms:
\lava~\cite{just2023lava}, \dataoob~\cite{kwon2023data},
\dvrl~\cite{yoon2020data}, and \knnshapley~\cite{jia2019efficient}.

For \sys, we set the Gaussian kernel bandwidth to the median of all pairwise distances and fix the balancing factor in Equation~\eqref{eq:net-if} to 0.03.
See details in Appendix~\ref{app:additional-exp}.

\vspace{-3mm}
\paragraph{Noise and Mislabel Detection.}
In this experiment, we introduce noise into 20\% of the data.
Following \cite{jiang2023opendataval,just2023lava,xie2017data,xie2018noising}, we inject feature noises by adding white noise to the images and randomly replacing words with other words for texts, and introduce label noises by randomly changing the labels of corrupted samples to other classes.

\begin{figure}[t]
    \centering
    \includegraphics[width=\textwidth]{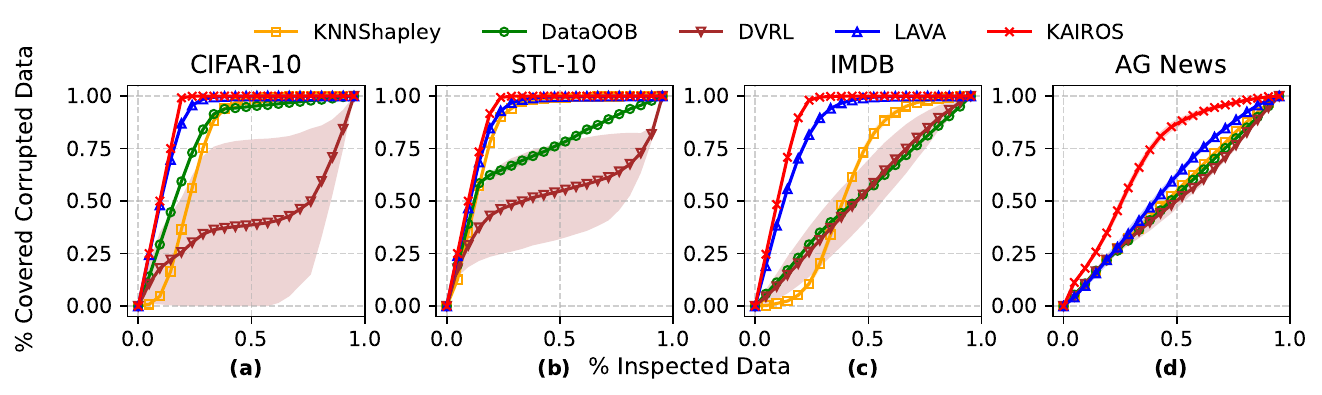}
    \vspace{-6mm}
    \caption{Feature noise detection results.}\vspace{-4mm}
    \label{fig:exp:noisy-detection}
\end{figure}

\begin{figure}[t]
    \centering
    \includegraphics[width=\textwidth]{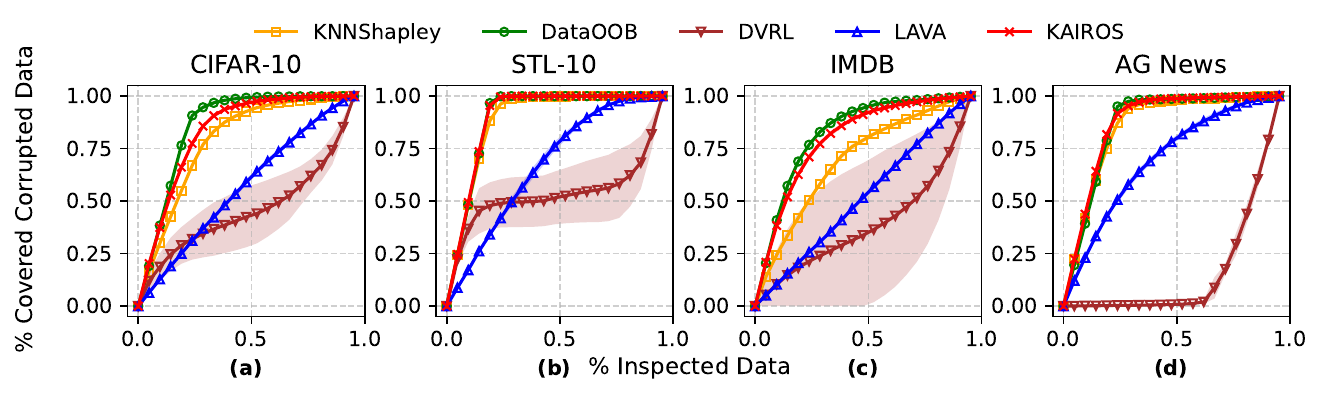}
    \vspace{-6mm}
    \caption{Label noise detection results.}\vspace{-5mm}
    \label{fig:exp:mislabel-detection}
\end{figure}

Figures~\ref{fig:exp:noisy-detection} and~\ref{fig:exp:mislabel-detection} present the performance of different data valuation methods in identifying corrupted samples stemming from two distinct sources of noise: feature perturbations and mislabels.
Each curve plots the cumulative fraction of corrupted data recovered as a function of the percentage of training data inspected.


\sys consistently achieves strong performance across both noise types and all datasets.
In the feature noise setting (Figure~\ref{fig:exp:noisy-detection}), it ranks noisy samples more effectively than all baselines on all datasets, especially in the early inspection phase.
In particular, on \dsagnews, all methods except \sys are close to the diagonal, meaning that they perform similarly to assigning random values to all data points.
In the presence of label noises (Figure~\ref{fig:exp:mislabel-detection}), our method remains competitive and outperforms most baselines, particularly on \dsagnews and \dsstl.

Overall, while some methods show competitive performance on specific datasets or noise types, such as \lava on feature noises and \dataoob on label noises, they fail to generalize well across both feature and label noise scenarios.
In contrast, \sys archives top-1 detection accuracy in 6 (out of 8) scenarios, and stays top-2 for the remaining 2 cases, indicating its versatile performance in noise detection.

\vspace{-3mm}
\paragraph{Malicious Data Detection.}
In this experiment, we evaluate robustness under adversarially crafted poisoning attacks.
For malicious data detection experiments, we follow \cite{just2023lava} to choose data poisoning attacks, including Badnet attack~\cite{gu2017badnets} and poison frogs attack~\cite{shafahi2018poison} on image data, i.e., CIFAR-10.

In addition, we choose the clean label style attack~\cite{qi2021mind} as a camouflaged poisoning attack for text data, where the triggers are not explicit words, but the style of the sentence.
We also test \textsc{Lisa}~\cite{huang2024lisa}, which aims to attack fine-tuning data for LLMs.

For Badnet and poison frog attacks, we choose the CIFAR-10 dataset, and inject 3\% malicious data.
For the style attack and \textsc{Lisa}, we choose the AG News dataset, and inject 10\% poisonous data.
Note that model-based techniques are not applicable to data for training generative models, as it is expensive to train a model hundreds of times on various bootstraps.

\begin{figure}[t]
    \centering
    \includegraphics[width=\textwidth]{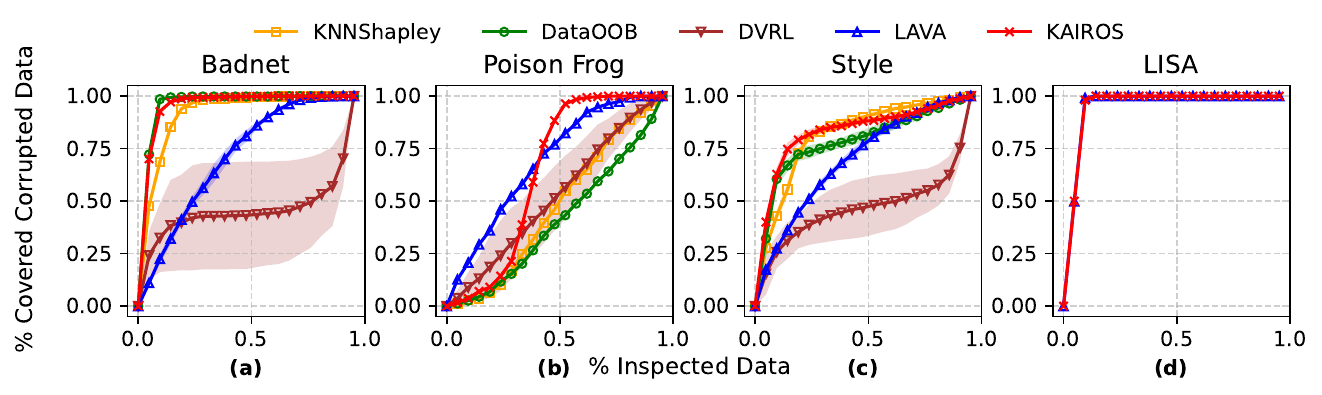}
    \vspace{-6mm}
    \caption{Malicious data detection results.}\vspace{-5mm}
    \label{fig:exp:corruption-detection}
\end{figure}

Across four scenarios shown in Figure~\ref{fig:exp:corruption-detection}, \lava falls short on (a) and (c), and \dataoob performs poorly on (b,c), and cannot adapt to (d).
Compared with \sys, \knnshapley, \dataoob, and \dvrl show less competitive performance across all cases. \lava achieves comparable performance with \sys in detecting poisoned fine-tuning data, and shows slightly better detection at the early stage of the poison frog attack detection.
However, it is significantly less effective in detecting the Badnet and style attacks.

\begin{figure}[t]
    \centering
    \includegraphics[width=\textwidth]{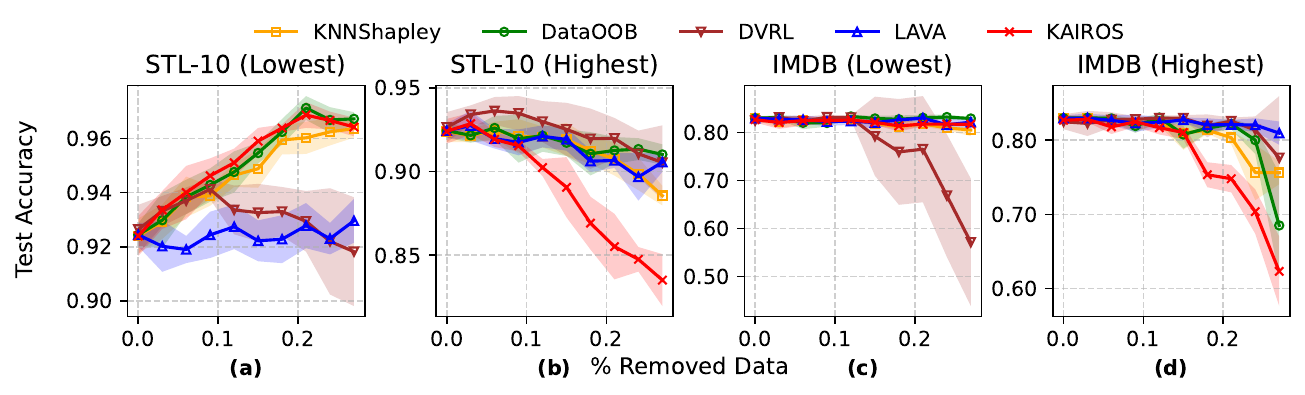}
    \vspace{-6mm}
    \caption{Effect of removing the least valuable (a,c) and the most valuable (b,d) data points on test accuracy.}\vspace{-5mm}
    \label{fig:exp:point-removal}
\end{figure}

\vspace{-3mm}
\paragraph{Point Removal.}
Figure~\ref{fig:exp:point-removal} (a–d) illustrates how test accuracy changes when either the least or most valuable training data, according to each method, is removed.
We test with 20\% of mislabels as it is more likely to affect the model's accuracy.
A high drop in accuracy when removing high-valued points and a low drop or increase in accuracy when removing low-valued points indicates good quality of valuations.

In both scenarios, our method yields the most desirable behavior.
When pruning the least-valuable data (Figure~\ref{fig:exp:point-removal} (a, c)), on \dsstl, \sys, \dataoob, and \knnshapley increase the test accuracy by a similar amount, while the values obtained from \dvrl and \lava do not help.
On \dsimdb data, all methods except \dvrl keep the test accuracy when removing 30\% of the data.
For both datasets, when discarding the most valuable data (Figure~\ref{fig:exp:point-removal} (b, d)), \sys results in more significant accuracy drops than all baselines.
This indicates that \sys gives both meaningful low and high values to the data, while most baselines only effectively identify low-valued data.

\begin{wrapfigure}[11]{r}{0.4\textwidth}
    \centering
    \vspace{-3mm}\includegraphics[width=.4\textwidth]{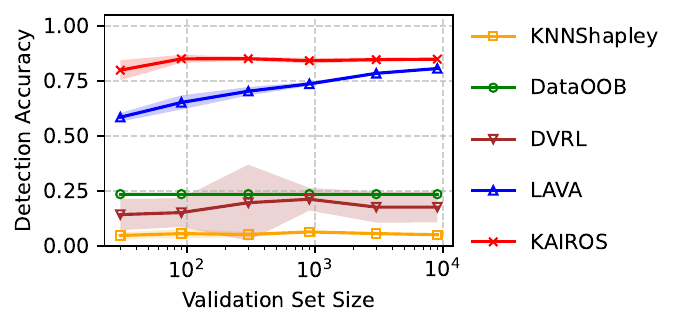}\vspace{-2mm}
    \caption{Label noise detection accuracy under varying validation set sizes.}
    \label{fig:exp:sample-size}
\end{wrapfigure}

\vspace{-3mm}
\paragraph{Effect of Validation Sample Size.}
In practice, validation sets are often small due to expensive labeling.
To understand how many validation samples are needed to obtain reliable data values, we conduct noise detection with varying validation set sizes.
In particular, to better compare the convergence of our method (based on MMD) and \lava (based on Wasserstein), we test with 20\% feature noises on \dsimdb data, the scenario where \lava performs relatively well.
To measure effectiveness, we adopt the detection accuracy, defined by the percentage of correctly identified corrupted data among the 20\% least-valued data points.
As shown in Figure~\ref{fig:exp:sample-size}, the variance of \sys, indicated by color regions, shrinks more quickly with growing validation set size than that of \lava.
Although \lava benefits from an increased validation set, it requires 9K validation samples to reach an accuracy of 0.77.
In contrast, \sys only requires 30 samples to achieve this.
This implies that MMD is more robust and reliable compared to Wasserstein under finite samples.

\begin{wrapfigure}[9]{r}{0.4\textwidth}
    \centering
    \vspace{-4mm}\includegraphics[width=.4\textwidth]{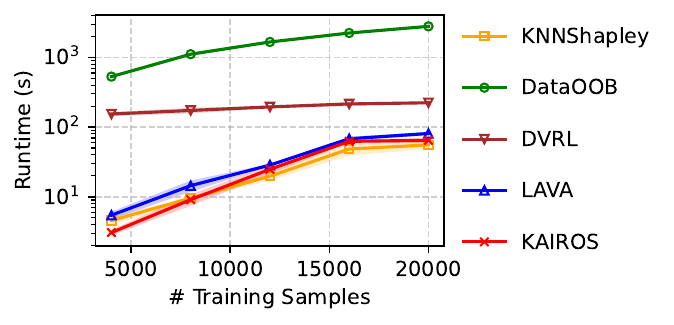}\vspace{-1.5mm}
    \caption{Offline runtime comparison.}
    \label{fig:exp:offline-runtime}
\end{wrapfigure}

\paragraph{Offline Runtime.}
To understand the scalability of different methods, we measure the runtime of methods on \dscifar data with label noise.
We vary the training set size and keep the static validation set size of 300.
As shown in Figure~\ref{fig:exp:offline-runtime}, \sys, \lava, and \knnshapley are similarly efficient, and are significantly faster than \dvrl (10x) and \dataoob (100x).
Although \dvrl's runtime grows slowly with \# training samples, as shown in previous results, it performs poorly in most applications, making it not practically applicable.
\dataoob's high runtime results from training models on various bootstraps.

\begin{wrapfigure}[12]{r}{0.4\textwidth}
    \centering

    \includegraphics[width=.4\textwidth]{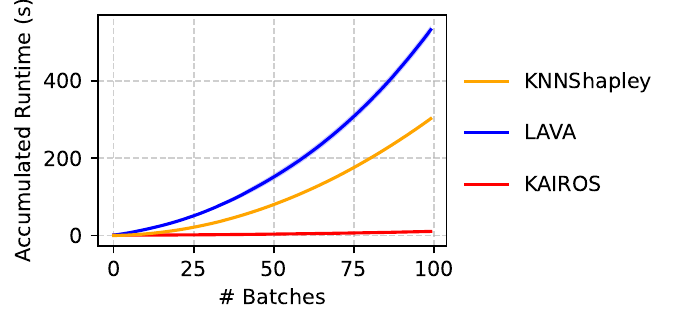}\vspace{-1.5mm}
    \caption{Online runtime comparison.}
    \label{fig:exp:online-runtime}
\end{wrapfigure}

\paragraph{Online Runtime.}
To show the adaptability of \sys in the online setting, we split 10000 of CIFAR-10 data into 100 batches, each containing 100 samples, and feed them in a streaming way.
We measure the accumulated time taken to conduct data valuation after each batch update.
\sys adopts Algorithm~\ref{alg:online-update} for value computation and updates, while \lava and \knnshapley have no direct adaptation to this setting, thus have to re-calculate the values when a new batch comes in.
\dataoob and \dvrl are omitted for this experiment as they take more than 8 hours to complete the experiment, meaning that they are not practical to use in this setting.
As shown in Figure~\ref{fig:exp:online-runtime}, \sys is significantly faster than \lava and \knnshapley in the online setting.
The speedup becomes more significant when more batches are included.
The speedup reaches 28x compared to \knnshapley, and 50x compared to \lava when all batches are fed.

\paragraph{Summary.}
Across all tasks and datasets, \sys consistently achieves performance gains over state-of-the-art baselines.
It effectively ranks data under both natural and adversarial data corruptions and noises.
The runtime experiments demonstrate the advantageous efficiency of \sys compared to baselines, especially in the practical online setting.

%% file: conclusions.tex
\vspace{-2mm}
\section{Conclusions, Limitations and Broader Impacts}\label{sec:conclusion}
\vspace{-2mm}
We introduce KAIROS, a scalable data valuation framework that uses Maximum Mean Discrepancy to compute closed-form influence functions for detecting feature noise, label corruption, and backdoors. KAIROS achieves up to 50$\times$ speedup over existing methods with $O(\batchsize N)$ complexity in online settings, making it practical for web-scale deployment while maintaining faithful leave-one-out rankings. Our approach provides theoretical guarantees through symmetry and density separation properties and offers model-agnostic influence scores that enable transparent data quality assessment, fairness auditing, and regulatory compliance without requiring model retraining. Current limitations include the use of fixed kernels and a fixed balancing coefficient for all tasks. Future work should focus on learned kernels, efficient approximate methods, and regression extensions.

%% file: appendix.tex
\newpage
\appendix
\input{literature_review}

\section{Proofs}
\subsection{Proof of Proposition~\ref{proposition:mmd_influence}}
\label{app:mmd_influence_proof}
\renewcommand{\theproposition}{\ref{proposition:mmd_influence}}
\begin{proposition}
The influence function for MMD as the distance metric is, up to additive and positive multiplicative constants, given by
\[
\mathrm{IF}_{\mathrm{MMD}}(x; P, Q) = \mathbb{E}_{x' \sim P}[k(x', x)] - \mathbb{E}_{x' \sim Q}[k(x', x)].
\]
\end{proposition}
\renewcommand{\theproposition}{\arabic{proposition}}

\begin{proof}
We first derive the influence function for $\MMD^2$, then apply the chain rule to obtain the result for $\MMD$.

\begin{align*}
\IF_{\MMD^2}(x; P, Q) &= -\left.\frac{d}{d\varepsilon} \MMD^2(P, (1-\varepsilon)Q + \varepsilon \delta_x) \right|_{\varepsilon = 0} \\
&= -\left.\frac{d}{d\varepsilon} \left\| \mu_P - \left((1-\varepsilon)\mu_Q + \varepsilon \phi(x)\right) \right\|_{\mathcal{H}}^2 \right|_{\varepsilon = 0} \\
&= -\left.\frac{d}{d\varepsilon} \left[ \|\mu_P - \mu_Q\|_{\mathcal{H}}^2 + 2\varepsilon \langle \mu_P - \mu_Q, \mu_Q - \phi(x) \rangle + \varepsilon^2 \|\mu_Q - \phi(x)\|^2 \right] \right|_{\varepsilon = 0} \\
&= -2 \langle \mu_P - \mu_Q, \mu_Q - \phi(x) \rangle \\
&= 2 \left( -\langle \mu_P, \mu_Q \rangle + \langle \mu_P, \phi(x) \rangle + \langle \mu_Q, \mu_Q \rangle - \langle \mu_Q, \phi(x) \rangle \right) \\
&= 2 \left( \mathbb{E}_{x' \sim P}[k(x', x)] - \mathbb{E}_{x' \sim Q}[k(x', x)]\right) \\
&- 2\left(\mathbb{E}_{x', x'' \sim P, Q}[k(x', x'')] + \mathbb{E}_{x', x'' \sim Q}[k(x', x'')] \right).
\end{align*}

Now, applying the chain rule:
\[
\IF_{\MMD}(x; P, Q) = \frac{\IF_{\MMD^2}(x; P, Q)}{2 \MMD(P, Q)}
\]
Ignoring terms independent of \(x\), we obtain the simplified expression:
\begin{align}
\IF_{\MMD}(x; P, Q) &= \mathbb{E}_{x' \sim P}[k(x', x)] - \mathbb{E}_{x' \sim Q}[k(x', x)]
\end{align}
\end{proof}

\subsection{Proof of Proposition~\ref{proposition:symmetry}}
\label{app:symmetry_proof}
\renewcommand{\theproposition}{\ref{proposition:symmetry}}
\begin{proposition}
Let $D^{\text{train}}$ and $D^{\text{val}}$ be finite samples from distributions $Q$ and $P$, respectively.
If for all subsets $S \subseteq D^{\text{train}} \setminus \{x_i^{\mathrm{train}}, x_j^{\mathrm{train}}\}$,
\[\widehat{\text{MMD}}(D^{\text{val}}, S \cup \{x_i^{\mathrm{train}}\}) - \widehat{\text{MMD}}(D^{\text{val}}, S) = \widehat{\text{MMD}}(D^{\text{val}}, S \cup \{x_j^{\mathrm{train}}\}) - \widehat{\text{MMD}}(D^{\text{val}}, S)\] then $\widehat{\mathrm{IF}}(x_i^{\mathrm{train}}) = \widehat{\mathrm{IF}}(x_j^{\mathrm{train}})$.
\end{proposition}
\renewcommand{\theproposition}{\arabic{proposition}}

\begin{proof}
Consider \(S = D^{\mathrm{train}} \setminus \{x_i^{\mathrm{train}},x_j^{\mathrm{train}}\}\).
We have,
\[
\widehat{\text{MMD}}\bigl(D^{\mathrm{val}}, S\cup\{x_i^{\mathrm{train}}\}\bigr)
- \widehat{\text{MMD}}\bigl(D^{\mathrm{val}}, S\bigr)
=
\widehat{\text{MMD}}\bigl(D^{\mathrm{val}}, S\cup\{x_j^{\mathrm{train}}\}\bigr)
- \widehat{\text{MMD}}\bigl(D^{\mathrm{val}}, S\bigr).
\]
Adding
\(\widehat{\text{MMD}}(D^{\mathrm{val}},S)\) to both sides gives
\[
\widehat{\text{MMD}}\bigl(D^{\mathrm{val}}, S\cup\{x_i^{\mathrm{train}}\}\bigr)
=
\widehat{\text{MMD}}\bigl(D^{\mathrm{val}}, S\cup\{x_j^{\mathrm{train}}\}\bigr).
\]
Squaring both sides,
\[
\widehat{\text{MMD}}^2\bigl(D^{\mathrm{val}}, S\cup\{x_i^{\mathrm{train}}\}\bigr)
=
\widehat{\text{MMD}}^2\bigl(D^{\mathrm{val}}, S\cup\{x_j^{\mathrm{train}}\}\bigr).
\]
Using the finite-sample estimator for \(\widehat{\text{MMD}}\),
\begin{align*}
\widehat{\mathrm{MMD}}^2(D^{\mathrm{val}}, T) &= \frac{1}{n_{\mathrm{val}}^2} \sum_{k, l} k(x_k^{\mathrm{val}},x_l^{\mathrm{val}}) + \frac{1}{|T|^2} \sum_{k, l\in T} k(x_k^{\mathrm{train}},x_l^{\mathrm{train}}) \\
&\quad - \frac{2}{n_{\mathrm{val}}\,|T|} \sum_{k=1}^{n_{\mathrm{val}}} \sum_{l\in T} k(x_k^{\mathrm{val}},x_l^{\mathrm{train}})
\end{align*}
where $T=S\cup\{x_i^{\mathrm{train}}\}$ or $T=S\cup\{x_j^{\mathrm{train}}\}$.
Therefore, substituting the finite-sample estimator into,
\[
\widehat{\text{MMD}}^2\bigl(D^{\mathrm{val}}, S\cup\{x_i^{\mathrm{train}}\}\bigr)
=
\widehat{\text{MMD}}^2\bigl(D^{\mathrm{val}}, S\cup\{x_j^{\mathrm{train}}\}\bigr).
\]
After canceling common terms and using $|T| = n_{\mathrm{train}}-1$,
\begin{align*}
&\frac{2}{n_{\mathrm{val}}} \sum_{m=1}^{n_{\mathrm{val}}} k(x_k^{\mathrm{val}}, x_i^{\mathrm{train}}) - \frac{2}{n_{\mathrm{train}}-1} \sum_{l = 1, l \neq i}^{n_{\mathrm{train}}} k(x_l^{\mathrm{train}}, x_i^{\mathrm{train}}) \\
&= \frac{2}{n_{\mathrm{val}}} \sum_{m=1}^{n_{\mathrm{val}}} k(x_k^{\mathrm{val}}, x_j^{\mathrm{train}}) - \frac{2}{n_{\mathrm{train}}-1} \sum_{l = 1, l \neq j}^{n_{\mathrm{train}}} k(x_l^{\mathrm{train}}, x_j^{\mathrm{train}})
\end{align*}
Therefore,
\[\widehat{\mathrm{IF}}(x_i^{\mathrm{train}}) = \widehat{\mathrm{IF}}(x_j^{\mathrm{train}})\]
\end{proof}

\subsection{Proof and Example of Proposition~\ref{proposition:threshold_separation}}
\label{app:threshold_separation_proof}
\paragraph{Intuition:} The core idea is that a threshold can be used to distinguish points where $P$ locally dominates $Q$ versus those where $Q$ dominates $P$, based on their relative densities in a neighborhood around each point. For accurate local density estimation, the kernel bandwidth $\sigma$ (e.g., in a Gaussian kernel) should be chosen sufficiently small. This result holds in the infinite sample regime where expectations are exact. In the finite sample case, this separation may not hold strictly, as small values of $\sigma$ can induce high-variance estimates.

\renewcommand{\theproposition}{\ref{proposition:threshold_separation}}
\begin{proposition}
Let $P$ and $Q$ be two probability distributions on $\mathcal{X}\subseteq\mathbb{R}^n$. For any $\epsilon>0$ and $r>0$, there exists a Gaussian isotropic kernel $k$ such that for \( \mathrm{IF}(x) \;=\;\mathbb{E}_{x'\sim P}\bigl[k(x,x')\bigr]\;-\;\mathbb{E}_{x'\sim Q}\bigl[k(x,x')\bigr]\),
the following holds
\[
\begin{cases}
\mathrm{IF}(x)>0, & \text{if }P(x')-Q(x')\ge\epsilon\ \forall\,x'\in B(x,r),\\
\mathrm{IF}(x)<0, & \text{if }P(x')-Q(x')\le-\epsilon\ \forall\,x'\in B(x,r),
\end{cases}
\]
where $B(x,r)=\{x':\|x'-x\|<r\}$.
\end{proposition}
\renewcommand{\theproposition}{\arabic{proposition}}

\begin{proof}
Let
\[
k(x,x')=\frac{1}{(2\pi\sigma^2)^{n/2}}
\exp\!\Bigl(-\frac{\|x-x'\|^2}{2\sigma^2}\Bigr),
\]
then
\[
\mathrm{IF}(x)
=\int_{x' \in \mathbb{R}^n}(P(x')-Q(x'))\,k(x,x')\,\mathrm{d}x'
=\int_{x' \in B(x,r)}(P-Q)\,k(x,x')
+\int_{x' \in B(x,r)^c}(P-Q)\,k(x,x').
\]
Since $P(x')-Q(x')\ge\epsilon$ for all $x' \in B(x,r)$ and $\lvert P-Q\rvert\le1$ outside,
\[
f(x)
\;\ge\;
\epsilon\! \int_{x' \in B(x,r)}k(x,x')
\;-\;\int_{x' \in B(x,r)^c}k(x,x')
=\epsilon A-(1-A)
=(\epsilon+1)A-1,
\]
where
\[
A=\int_{x' \in B(x,r)}k(x,x')\,\mathrm{d}x'
\]

Note that,
\[
\int_{B(x,r)}k(x,x')\,\mathrm{d}x'
=\Pr_{X\sim\mathcal N(0,\sigma^2I)}(\|X\|<r)
=\Pr(\|X\|^2<r^2)
=1-\Pr(\|X\|^2\ge r^2).
\]

Let $W=\|X\|^2=\sum_{i=1}^nX_i^2$, so
$\mathbb{E}[W]=n\sigma^2$ and $\text{Var}(W)=2n\sigma^4$. By Chebyshev's inequality, for $r^2>n\sigma^2$,
\[
\Pr(W\ge r^2)
\le\frac{\text{Var}(W)}{(r^2-\mathbb{E}[W])^2}
=\frac{2n\sigma^4}{(r^2-n\sigma^2)^2},
\]
hence
\[
A
\ge1-\frac{2n\sigma^4}{(r^2-n\sigma^2)^2}.
\]

Choose $\frac{r}{\sqrt{n}} > \sigma^*>0$ so that
\[
1-\frac{2n{\sigma^*}^4}{(r^2-n{\sigma^*}^2)^2}
>\frac{1}{\epsilon+1}.
\]

Then $(\epsilon+1)A-1>0$, implying $\mathrm{IF}(x)>0$.

The same $\sigma^*$ gives
\[
\mathrm{IF}(x)\le -\bigl((\epsilon+1)A-1\bigr)<0
\]
for the other case where $P(x')-Q(x')\le-\epsilon$.

Thus, there exists a Gaussian kernel with bandwidth $\sigma^*$ that satisfies the desired separation.
\end{proof}

\begin{figure}
  \centering
  \vspace{-4mm}
  \includegraphics[width=.8\linewidth]{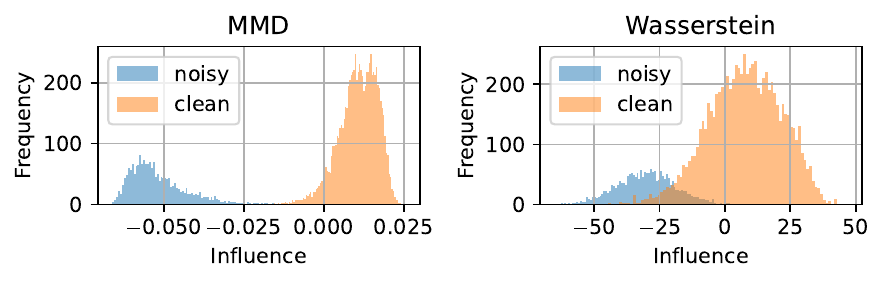}\vspace{-2mm}
  \caption{Influence distribution obtained from influence based on MMD (\sys) and Wasserstein (\lava).}
  \label{fig:separation}
\end{figure}
\begin{example}[Density Separation]
We compare the influence distributions obtained from MMD and Wasserstein, computed on the CIFAR-10 sample data, to show the density separation property.
As shown in Figure~\ref{fig:separation}, MMD-based influence exhibits a near-zero optimal threshold that separates the noisy and clean points almost perfectly. On the other hand, the Wasserstein-based influence for noisy and clean data is entangled, and there is no threshold which cleanly separates the two.

\end{example}

\subsection{Proof of Proposition~\ref{proposition:emcmd_influence}}
\label{app:emcmd-influence-proof}
\renewcommand{\theproposition}{\ref{proposition:emcmd_influence}}
\begin{proposition}
The influence function for \EMCMD, up to additive and positive multiplicative constants, given by
\begin{equation}
\mathrm{IF}_{\EMCMD}((x,y); P, Q) = \|\mu_{Y|X}^P(x) - \phi(y)\|_{\mathcal{H}_\mathcal{Y}}
\end{equation}
\end{proposition}
\renewcommand{\theproposition}{\arabic{proposition}}
\begin{proof}
Consider the perturbed distribution
\(
Q_\varepsilon = (1-\varepsilon)Q + \varepsilon \delta_{(x,y)},
\)
with corresponding marginal
\(
(1-\varepsilon)Q_X+\varepsilon\delta_{x}
\). Since $Q_\varepsilon(Y=y\mid x) = 1$, the perturbed conditional embedding at point \( x\) is given by
\[
\mu_{Y|X}^{Q_\varepsilon}(x)=\phi(y),
\]

Therefore,
\begin{equation}
\MCMD_{P,Q_\varepsilon}(x') =
\begin{cases}
\MCMD_{P,Q}(x') & \text{if } x' \neq x \\
\|\mu_{Y|X}^P(x) - \phi(y)\|_{\mathcal{H}_\mathcal{Y}} & \text{if } x' = x
\end{cases}
\end{equation}

\(\EMCMD(P,Q_\varepsilon)\) can be written as:
\begin{flalign*}\small
&\mathbb{E}_{X\sim(1-\varepsilon)Q_X + \varepsilon \delta_{x}}[\mathrm{MCMD}_{P,(1-\varepsilon)Q + \varepsilon \delta_{(x,y)}}(X)] && \\
&=(1-\varepsilon) \left(\mathbb{E}_{X\sim Q_X}[\mathrm{MCMD}_{P,Q_\varepsilon}(X)]\right) + \varepsilon \left(\mathrm{MCMD}_{P,Q_\varepsilon}(x)\right) && \\
&=(1-\varepsilon) \left(\mathbb{E}_{X\sim Q_X}[\mathrm{MCMD}_{P,Q}(X)]\right) + \varepsilon \|\mu_{Y|X}^P(x) - \phi(y)\|_{\mathcal{H}_\mathcal{Y}} &&
\end{flalign*}
Therefore we have,
\begin{align*}
&&&\mathrm{IF}_{\EMCMD}((x,y); P, Q) \\
&&=& -\lim_{\varepsilon \rightarrow 0^+} \frac{\mathbb{E}_{X \sim (1-\varepsilon)Q_X + \varepsilon \delta_{x}}[\mathrm{MCMD}_{P,(1-\varepsilon)Q + \varepsilon \delta_{x}}(X)] - \mathbb{E}_{X \sim Q_X}[\mathrm{MCMD}_{P,Q}(X)]}{\varepsilon}\\
&&=& -\lim_{\varepsilon \rightarrow 0^+} \frac{(1-\varepsilon) \left(\mathbb{E}_{X\sim Q_X}[\mathrm{MCMD}_{P,Q}(X)]\right) + \varepsilon \|\mu_{Y|X}^P(x) - \phi(y)\|_{\mathcal{H}_\mathcal{Y}} - \mathbb{E}_{X \sim Q_X}[\mathrm{MCMD}_{P,Q}(X)]}{\varepsilon}\\
&&=& -\lim_{\varepsilon \rightarrow 0^+} \frac{\varepsilon \|\mu_{Y|X}^P(x) - \phi(y)\|_{\mathcal{H}_\mathcal{Y}} - \varepsilon\mathbb{E}_{X \sim Q_X}[\mathrm{MCMD}_{P,Q}(X)]}{\varepsilon}\\
&&=& -\|\mu_{Y|X}^P(x) - \phi(y)\|_{\mathcal{H}_\mathcal{Y}} + \mathbb{E}_{X \sim Q_X}[\mathrm{MCMD}_{P,Q}(X)]
\end{align*}

Ignoring terms independent of \(x\) and \(y\),
\[\mathrm{IF}_{\EMCMD}((x,y); P, Q) = -\|\mu_{Y|X}^P(x) - \phi(y)\|_{\mathcal{H}_\mathcal{Y}}\]

\end{proof}

\subsection{Proof of Theorem~\ref{theorem:generalization-bound}}\label{app:generalization-bound-proof}
We present the full statement of Theorem~\ref{theorem:generalization-bound} below, followed by a detailed proof.
\renewcommand{\thetheorem}{\ref{theorem:generalization-bound}}
\begin{theorem}[Bounding transfer loss]
Let $(\mathcal{X},d_{\mathcal{X}})$ and $(\mathcal{Y},d_{\mathcal{Y}})$ be compact metric spaces, and let $k_{\mathcal{X}},k_{\mathcal{Y}}$ be universal kernels on $\mathcal{X},\mathcal{Y}$ with RKHS $\mathcal{H}_{\mathcal{X}},\mathcal{H}_{\mathcal{Y}}$.  Equip $\mathcal{Z}=\mathcal{X}\times\mathcal{Y}$ with the tensor‐product kernel whose RKHS is
\[
\mathcal{H} = \mathcal{H}_{\mathcal{X}}\widehat\otimes\mathcal{H}_{\mathcal{Y}}.
\]
Let $P,Q\in\mathcal{P}(\mathcal{Z})$ have marginals $P_{X},Q_{X}\in\mathcal{P}(\mathcal{X})$ and conditionals $P(\cdot\mid x),Q(\cdot\mid x)\in\mathcal{P}(\mathcal{Y})$.  Let $L:\mathcal{Z}\to\mathbb{R}$ be a continuous loss function. Since \(\mathcal{H}\) is endowed with a universal kernel, $\mathcal{H}$ is dense in $C(\mathcal{Z})$. Let
\[
\|L\|_* =
\inf_{L'\in\mathcal{H}}\Bigl\{\,
\big\|\mathbb{E}_{y\sim P(\cdot\mid x)}L'(x,y)\big\|_{\mathcal{H}_{\mathcal{X}}}
+
\sup_{x\in\mathcal{X}}\big\| \sup_{y\in\mathcal{Y}}L'(x,y)\big\|_{\mathcal{H}_{\mathcal{Y}}}
\;+\;\|L-L'\|_\infty
\Bigr\}.
\]
Then
\[
\mathbb{E}_{(x,y)\sim Q}[L(x,y)
] \leq
\mathbb{E}_{(x,y)\sim P}[L(x,y)] +
\|L\|_*
\Bigl(\,
\mathrm{MMD}_{\mathcal{X}}(P_{X},Q_{X})
+
\mathbb{E}_{x\sim Q_{X}}\bigl[\mathrm{MCMD}_{P,Q}(x)\bigr]
\Bigr).
\]
\end{theorem}
\renewcommand{\thetheorem}{\arabic{theorem}}

\begin{proof}
Fix $\varepsilon>0$. Since $\mathcal{H}$ is dense in $C(\mathcal{Z})$, choose $L'\in\mathcal{H}$ with $\|L-L'\|_{\infty}\le\varepsilon$. Then
\[
\mathbb{E}_{Q}[L]
= \mathbb{E}_{Q}[L'] + \mathbb{E}_{Q}[L-L'],
\qquad
\mathbb{E}_{P}[L]
= \mathbb{E}_{P}[L'] + \mathbb{E}_{P}[L-L'],
\]
so
\[
\mathbb{E}_{Q}[L] - \mathbb{E}_{P}[L]
= \bigl(\mathbb{E}_{Q}[L'] - \mathbb{E}_{P}[L']\bigr)
+ \bigl(\mathbb{E}_{Q}[L-L'] - \mathbb{E}_{P}[L-L']\bigr).
\]
Since $P$ and $Q$ are probability measures,
\[
\bigl|\mathbb{E}_{Q}[L-L'] - \mathbb{E}_{P}[L-L']\bigr|
\le \|L-L'\|_{\infty}
\le \varepsilon.
\]
By the law of total expectation,
\[
\mathbb{E}_{Q}[L']
= \mathbb{E}_{x\sim Q_{X}}\bigl[\mathbb{E}_{y\sim Q(\cdot\mid x)}L'(x,y)\bigr],
\quad
\mathbb{E}_{P}[L']
= \mathbb{E}_{x\sim P_{X}}\bigl[\mathbb{E}_{y\sim P(\cdot\mid x)}L'(x,y)\bigr].
\]
Add and subtract $\mathbb{E}_{x\sim Q_{X}}\mathbb{E}_{y\sim P(\cdot\mid x)}L'$ to obtain
\[
\mathbb{E}_{Q}[L'] - \mathbb{E}_{P}[L']
= A + B,
\]
where
\[
A
= \mathbb{E}_{x\sim Q_{X}}\Bigl[\mathbb{E}_{y\sim Q(\cdot\mid x)}L'(x,y)
- \mathbb{E}_{y\sim P(\cdot\mid x)}L'(x,y)\Bigr],
\]
\[
B
= \mathbb{E}_{x\sim Q_{X}}\Bigl[\mathbb{E}_{y\sim P(\cdot\mid x)}L'(x,y)\Bigr]
- \mathbb{E}_{x\sim P_{X}}\Bigl[\mathbb{E}_{y\sim P(\cdot\mid x)}L'(x,y)\Bigr].
\]
Define $f_{L'}(x)=\mathbb{E}_{y\sim P(\cdot\mid x)}L'(x,y)\in\mathcal{H}_{\mathcal{X}}$. Then
\[
B
= \mathbb{E}_{x\sim Q_{X}}f_{L'}(x) - \mathbb{E}_{x\sim P_{X}}f_{L'}(x),
\]
so
\[
|B|
\le \|f_{L'}\|_{\mathcal{H}_{\mathcal{X}}}\,\mathrm{MMD}_{\mathcal{X}}(P_{X},Q_{X})
= \|L'\|_{\mathcal{H}}^{(\mathcal{X})}\,\mathrm{MMD}_{\mathcal{X}}(P_{X},Q_{X}).
\]
For each $x$, let $L'_{y}(x,\cdot)\in\mathcal{H}_{\mathcal{Y}}$. Then
\[
\bigl|\mathbb{E}_{y\sim Q(\cdot\mid x)}L'(x,y) - \mathbb{E}_{y\sim P(\cdot\mid x)}L'(x,y)\bigr|
\le \|L'_{y}(x,\cdot)\|_{\mathcal{H}_{\mathcal{Y}}}\,\mathrm{MCMD}_{P,Q}(x),
\]
hence
\[
|A|
\le \mathbb{E}_{x\sim Q_{X}}\bigl[\|L'\|_{\mathcal{H}}^{(\mathcal{Y})}\,\mathrm{MCMD}_{P,Q}(x)\bigr]
= \|L'\|_{\mathcal{H}}^{(\mathcal{Y})}\,\mathbb{E}_{x\sim Q_{X}}[\mathrm{MCMD}_{P,Q}(x)].
\]
Combining the bounds for $A$, $B$, and the approximation term,
\[
\mathbb{E}_{Q}[L]
\le \mathbb{E}_{P}[L]
+ \|L'\|_{\mathcal{H}}^{(\mathcal{X})}\,\mathrm{MMD}_{\mathcal{X}}(P_{X},Q_{X})
+ \|L'\|_{\mathcal{H}}^{(\mathcal{Y})}\,\mathbb{E}_{x\sim Q_{X}}[\mathrm{MCMD}_{P,Q}(x)]
+ \varepsilon.
\]
Taking the infimum over $L'\in\mathcal{H}$ yields the desired result.
\end{proof}

\section{Influence Computation and Update for Streaming data}
\label{appendix:algorithm-online}
Algorithm~\ref{alg:online-update} provides the detailed algorithm for computing the influence in the online setting.

\begin{algorithm}[H] \small
\caption{Online update for influence}
\label{alg:online-update}
\begin{algorithmic}[1]
\Require
  Current size \(n_t\), variables \(\AvgTrainK^{(t)},\AvgValK^{(t)},\Residual^{(t)}\in\mathbb{R}^{n_t}\), validation size \(n_{\mathrm{val}}\), batch size \(\batchsize\), incoming batch data \((X^{\mathrm{new}}\in\mathbb{R}^{\batchsize\times d},\,Y^{\mathrm{new}}\in\mathbb{R}^{\batchsize\times c})\), balancing factor \(\lambda\).
\Ensure
  Updated \(\AvgTrainK^{(t+1)},\AvgValK^{(t+1)},\Residual^{(t+1)},\FinalScore^{(t+1)}\in\mathbb{R}^{n_{t+1}}\).

\State \(n_{t+1}\gets n_t + \batchsize\)
\State compute
  \(K^{\mathrm{old,new}}\in\mathbb{R}^{n_t\times \batchsize},\;
   K^{\mathrm{new,old}}\in\mathbb{R}^{\batchsize\times n_t},\;
   K^{\mathrm{new,new}}\in\mathbb{R}^{\batchsize\times \batchsize}\)
  \hfill {\color{blue}// kernel sub‐matrices}
\For{\(i=1,\dots,n_t\)} \hfill {\color{blue}// update kernel means for old points}
  \State \(\AvgTrainK_i^{(t+1)}
    \gets \tfrac{1}{n_{t+1}-1} \big((n_t - 1)\,\AvgTrainK_i^{(t)}
    + \sum_{j=1}^{\batchsize}K^{\mathrm{old,new}}_{i,j}\big)\)
  \State \(\AvgValK_i^{(t+1)}\gets \AvgValK_i^{(t)}\)
  \State \(\Residual_i^{(t+1)}\gets \Residual_i^{(t)}\)
\EndFor

\State \(\Residual^{\mathrm{new}}_{i}
       = \bigl\|y_{n_t+i}^{\mathrm{train}}
       - \hat y_{n_t+i}^{\mathrm{train}}\bigr\|
       \quad(i=1,\dots,\batchsize)\)
  \hfill {\color{blue}// residuals for new points}
\For{\(i=1,\dots,\batchsize\)} \hfill {\color{blue}// update kernel means for new points}
  \State \(\AvgTrainK_{n_t+i}^{(t+1)}
    \gets \tfrac{1}{n_{t+1}-1} \big(\sum_{j=1}^{n_t}K^{\mathrm{new,old}}_{i,j}
      + \sum_{j'=1}^{\batchsize}K^{\mathrm{new,new}}_{i,j'}
      - 1\big)\)
  \State \(\AvgValK_{n_t+i}^{(t+1)}
    \gets \tfrac{1}{n_{\mathrm{val}}}
      \sum_{j=1}^{n_{\mathrm{val}}}
      k\bigl(x_j^{\mathrm{val}},x_{n_t+i}^{\mathrm{train}}\bigr)\)
  \State \(\Residual_{n_t+i}^{(t+1)}
    \gets \Residual^{\mathrm{new}}_{i}\)
\EndFor

\For{\(i=1,\dots,n_{t+1}\)}
  \State \(\FinalScore_i^{(t+1)}
    \gets \lambda\bigl(\AvgValK_i^{(t+1)}-\AvgTrainK_i^{(t+1)}\bigr)
      + (1-\lambda)\,\Residual_i^{(t+1)}\)
  \hfill {\color{blue}// net influence}
\EndFor
\end{algorithmic}
\end{algorithm}

\section{Additional Experiments and Detailed Settings}\label{app:additional-exp}

\paragraph{Detailed description of baselines and choice of hyper-parameters.}
We compare with the following baselines:
\lava~\cite{just2023lava} is a model‐agnostic method that uses the Wasserstein-1 dual potential as an influence proxy;
\dataoob~\cite{kwon2023data} trains a bootstrap ensemble and assigns each point its average out-of-bag loss to capture its contribution;
\dvrl~\cite{yoon2020data} learns per‐sample importance weights via reinforcement learning and uses those weights directly as data scores;
and \knnshapley~\cite{jia2019efficient} computes exact Shapley values on a nearest‐neighbor proxy model, exploiting its closed‐form solution for efficiency.
To avoid the prohibitive runtime of \dataoob without sacrificing accuracy, we decrease its number of bootstraps from 1000 to 100.
All baselines run with their default settings in OpenDataVal~\cite{jiang2023opendataval}.

Following \cite{garreau2017large,biggs2023mmd,gretton2012kernel}, we adopt the "median heuristic" to set the kernel bandwidth to the median of pair-wise distances.
Given the large number of samples, we estimate the median on 10000 sampled pairs.
In practice, as shown in experiments, sampled bandwidths work well across various datasets and scenarios.
The balancing factor is determined by aligning the scale of the two components of net influence (Equation~\eqref{eq:net-if}), when both feature and label noise exist.

\paragraph{Dataset Licensing Information.}
The license terms for each dataset used in this work are as follows:
\begin{itemize}[leftmargin=*]
  \item \dscifar \cite{krizhevsky2009learning}: released under the MIT License.
        \url{https://www.cs.toronto.edu/~kriz/cifar.html}
  \item \dsstl \cite{coates2011analysis}: no explicit license is provided.
        \url{http://ai.stanford.edu/~acoates/stl10/}
  \item \dsimdb \cite{maas2011learning}: subject to IMDb’s non-commercial terms of use.
        \url{https://datasets.imdbws.com/}
  \item \dsagnews \cite{zhang2015character}: provided for research and non-commercial use.
        \url{http://www.di.unipi.it/~gulli/AG_corpus_of_news_articles.html}
\end{itemize}




%% file: literature_review.tex
\section{Related Work}
\label{section:related_work}
\paragraph{Model-based Data Valuation}
A common approach for data valuation is the leave-one-out (LOO) score, i.e., the change in performance when a point is removed from training, but this is computationally expensive. Influence functions~\citep{koh2017understanding} approximate these scores using second-order derivatives, though they remain intractable for modern deep models. Recent advances have made influence functions scalable to large models~\citep{grosse2023studying, choe2024your}, and others trace test loss over the training trajectory to attribute influence~\citep{pruthi2020tracin, bae2024training, wang2024data}. \textsc{Trak}~\citep{park2023trak} takes a post-hoc approach, approximating the model as a kernel machine to trace predictions back to training examples, while \textsc{DaVinz}~\citep{wu2022davinz} uses neural tangent kernel approximations to predict influence at initialization. Other approaches include reinforcement learning to learn data values~\citep{yoon2020data}, prediction from noisy labels~\citep{covert2024stochastic}, and dynamic self-weighting mechanisms within loss functions~\citep{wibiral2024lossval}.

\paragraph{Algorithm-based Data Valuation}

Unlike model-based methods that track influence during or after training, algorithm-based approaches define data value in terms of a specific learning algorithm and utility function. A major line relies on Shapley values~\citep{shapley1953value}, which uniquely satisfy fairness axioms such as symmetry and null-value. Data Shapley~\citep{ghorbani2019data} defines a point’s value as its average marginal contribution to model utility across all subsets, but exact computation requires training \(2^N\) models. Efficient approximations~\citep{jia2019towards}, closed-form solutions for \(k\)-NN~\citep{jia2019efficient}, class-wise extensions~\citep{schoch2022cs}, and distributional variants~\citep{ghorbani2020distributional} have been proposed. Retraining multiple models can be avoided through gradient- and hessian-based approximations~\citep{wang2024data}. However, the reliability of shapley methods is sensitive to the utility function~\citep{wang2024rethinking}. Other work relaxes the efficiency axiom to improve robustness~\citep{kwon2021beta, wang2023data}. For bagging models, Kwon et al.~\cite{kwon2023data} show that out-of-bag estimates yield effective approximations.

\paragraph{Algorithm-agnostic Data Valuation}
Algorithm-agnostic valuation methods operate without knowledge of the learning algorithm. Just et al.~\citep{just2023lava} propose \lava, which quantifies the contribution of a point based on its contribution to the dataset distance. However, their method has several limitations: (i) it performs poorly in identifying label errors~\citep{jiang2023opendataval}; (ii) their approximations are not always aligned with the leave one out valuation and can be indeterministic~\citep{villani2008optimal} (see Figure~\ref{fig:intro:sinkhorn-dual}); and (iii) the method is computationally expensive (even in approximate form, has \(O(n^2)\) complexity)~\citep{cuturi2013sinkhorn}. The distance must be computed over the entire dataset even when valuing a single point (iv)the approximate form is not a distance measure (v)the memory complexity is \(O(n^2)\) . To address memory bottlenecks in \textsc{Lava} for large datasets, Kessler et al.~\citep{kessler2024sava} propose \textsc{Sava}, a batching strategy for Wasserstein computation, though other challenges remain unresolved.

\paragraph{Dataset Valuation}
Another line of work focuses on valuing entire datasets rather than individual points, typically in settings with multiple data providers where fair compensation is desired~\citep{chen2020truthful, amiri2023fundamentals}. These methods use distance-based metrics such as mutual information~\citep{chen2020truthful}, MMD\(^2\)~\citep{tay2022incentivizing}, MMD~\citep{xu2024data}, and volume~\citep{xu2021validation} to assess utility. However, they do not extend naturally to individual valuation due to their reliance on large datasets, inability to capture point-level interactions, and lack of influence estimation.